\documentclass[12pt]{article}
\usepackage{amsmath}
\usepackage{graphicx}
\usepackage{natbib}
\usepackage{diagbox}
\usepackage{url} 

\newcommand{\blind}{0}

\begin{document}

\def\spacingset#1{\renewcommand{\baselinestretch}%
{#1}\small\normalsize} \spacingset{1}


\if0\blind
{
  \title{\bf Performance is not enough:\\ the story told by a Rashomon quartet}
  \author{Przemysław Biecek
    \\
    {\small MI2.AI, Warsaw University of Technology \& University of Warsaw}\\
    and \\
    Hubert Baniecki \\
    {\small MI2.AI, University of Warsaw}\\
    and \\
    Mateusz Krzyziński \\
    {\small MI2.AI, Warsaw University of Technology}\\
    and \\
    Dianne Cook \\
    {\small Monash University}
    }
  \maketitle
} \fi

\if1\blind
{
  \bigskip
  \bigskip
  \bigskip
  \begin{center}
    {\LARGE\bf Performance is not enough:\\ the story told by a Rashomon Quartet}
\end{center}
  \medskip
} \fi

\bigskip
\begin{abstract}
The usual goal of supervised learning is to find the best model, the one that optimizes a particular performance measure. However,  what if the explanation provided by this model is completely different from another model and different again from another model despite all having similarly good fit statistics? Is it possible that the equally effective models put the spotlight on different relationships in the data?
Inspired by \emph{Anscombe's quartet}, this paper introduces a \emph{Rashomon Quartet}, i.e. a set of four models built on a synthetic dataset which have practically identical predictive performance. However, the visual exploration reveals distinct explanations of the relations in the data. This illustrative example aims to encourage the use of methods for model visualization to compare predictive models beyond their performance. 
\end{abstract}

\noindent%
\vfill

\newpage 
\spacingset{1.5} 

{\large\bfseries  Fifty years ago}, Francis Anscombe published a groundbreaking paper ``Graphs in Statistical Analysis''~\citep{Anscombe73}, in which he argued that visualization is a crucial element for statistical analysis. He introduced a set of four datasets, later called \emph{Anscombe's quartet}, which are described by identical statistical characteristics and nevertheless express completely different relationships between the variables. This was a simple but very powerful argument towards visualizing datasets to better understand relations between variables. Datasets that, through the lens of numerical summaries, appear to be similar, may in fact look very differently.

This is also true for predictive models. As we will see, models of different classes trained to predict the same response may achieve identical $R^2$ and $RMSE$ performance values but explain the data very differently. 

\begin{figure}[!htb]
\begin{center}
\includegraphics[width=0.8\textwidth]{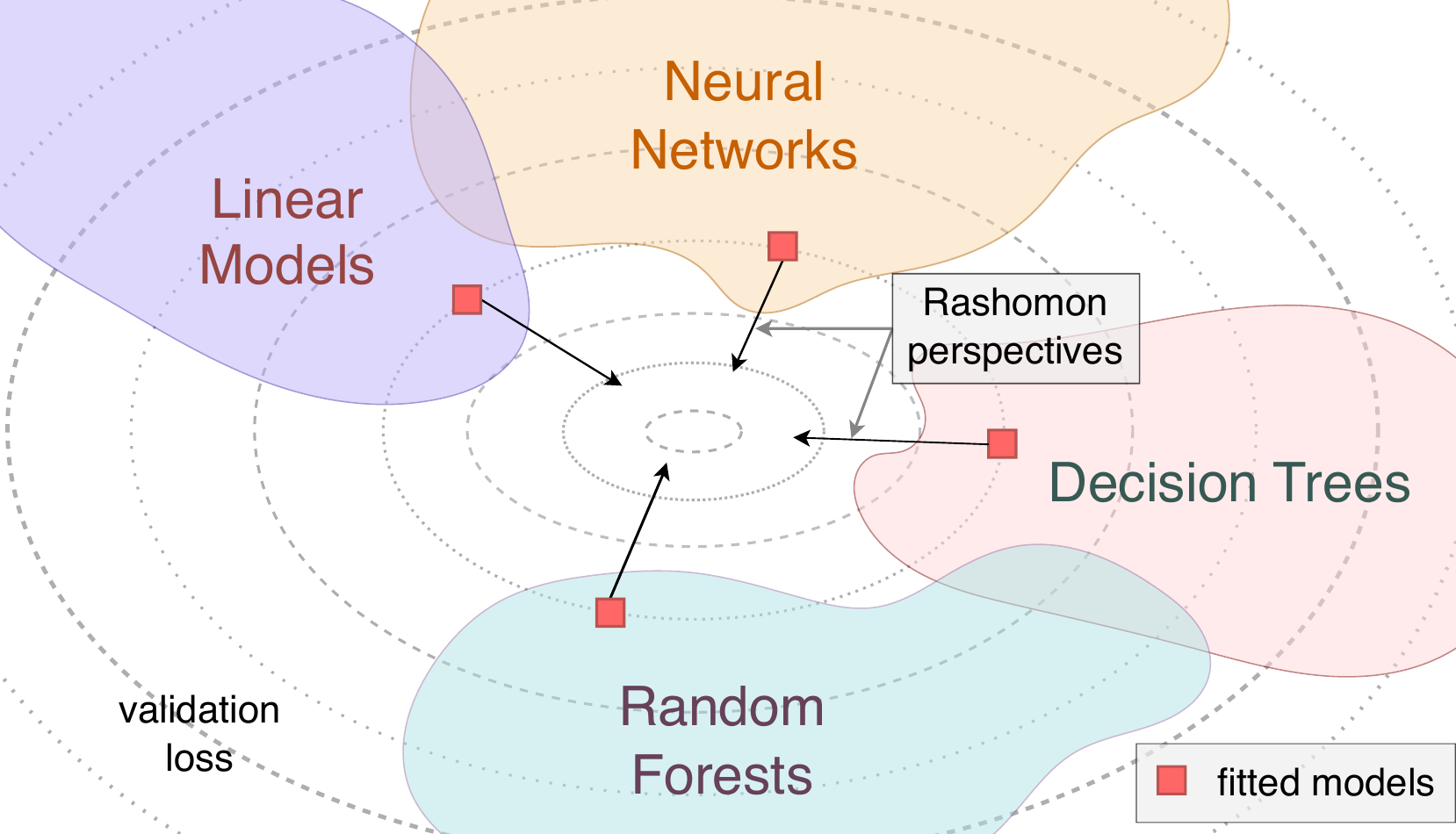}
\end{center}
\caption{Illustration of the Rashomon effect: equally effective models, each telling a different story. Dashed contours indicate an equal value of a loss function calculated on a validation dataset. Squares indicate ``best in class'' models trained on the same data. The \emph{Rashomon Quartet} is designed so that the best models have an equal loss function on the validation data, but each model from the set describes a different perspective.}\label{fig:diagram-rashomon}
\end{figure}

Similar performance of the best-fitted models does not mean that they encode similar stories about data. Leo Breiman highlighted this in his seminal paper ``Statistical Modeling: The Two Cultures''~\citep{twoCultures}, in which he introduced the term \emph{Rashomon effect} to describe this phenomenon. The name ``Rashomon'' refers to the title of a movie by Akira Kurosawa, in which four witnesses describe the same event in very different ways. American artist Chuck Ginnever also adapted the idea in his work named Rashomon, where 15 identical pieces have 15 different orientations that make them appear to be different~\citep{ginnever}.  As an illustration of this concept, Figure \ref{fig:diagram-rashomon} shows the hypothetical value of a loss function in the space of all possible predictive models. The Rashomon effect means that multiple ``best'' models found in one family of models or even different families can have the same value of the loss function. Statisticians know perfectly well, paraphrasing \cite{Box}, that ``all models are wrong", but how then do you know which ones ``are useful" when the summary of the fits are identical?

Today, the \emph{Rashomon effect} is being studied from various perspectives. For example, \citet{MCRRudin} introduces \emph{Rashomon set} to define the importance of a variable not for a single model but for an entire class of predictive models. With linear models, the Rashomon set can be studied analytically, but in the general case, proper characterization of such a set is a challenge yet to be solved~\citep{RudinChallenges}. \citet{Kasia} uses \emph{Rashomon set} to segment machine learning models with respect to their behavior, allowing for better analysis of the data-generating process. Complex models, such as tree ensembles and neural networks, are relatively difficult to characterize due to having a large number of parameters. Model visualization techniques are valuable alternatives to performance measures, both focusing on the statistical perspective~\citep{WickhamCH15}, as well as the predictive one~\citep[e.g.][]{ema2021}. While conventionally, \emph{Rashomon effect} is used to refer to analyzing diverging models, \citet{baniecki2023grammar} adheres to it in juxtaposing complementary explanations for a more comprehensive visual model analysis.
In deep learning, where model complexity increases exponentially, an aggregation of explanations can be used to quantify consistency between the model set~\citep[see e.g.][]{watson2022agree}. Finally, \citet{gelman2023causal} introduce \emph{Causal quartets} to show that a single value of average causal effect may represent very different causal patterns, even on simple synthetic data.

\begin{figure}[!p]
\begin{center}
\includegraphics[width=0.95\textwidth]{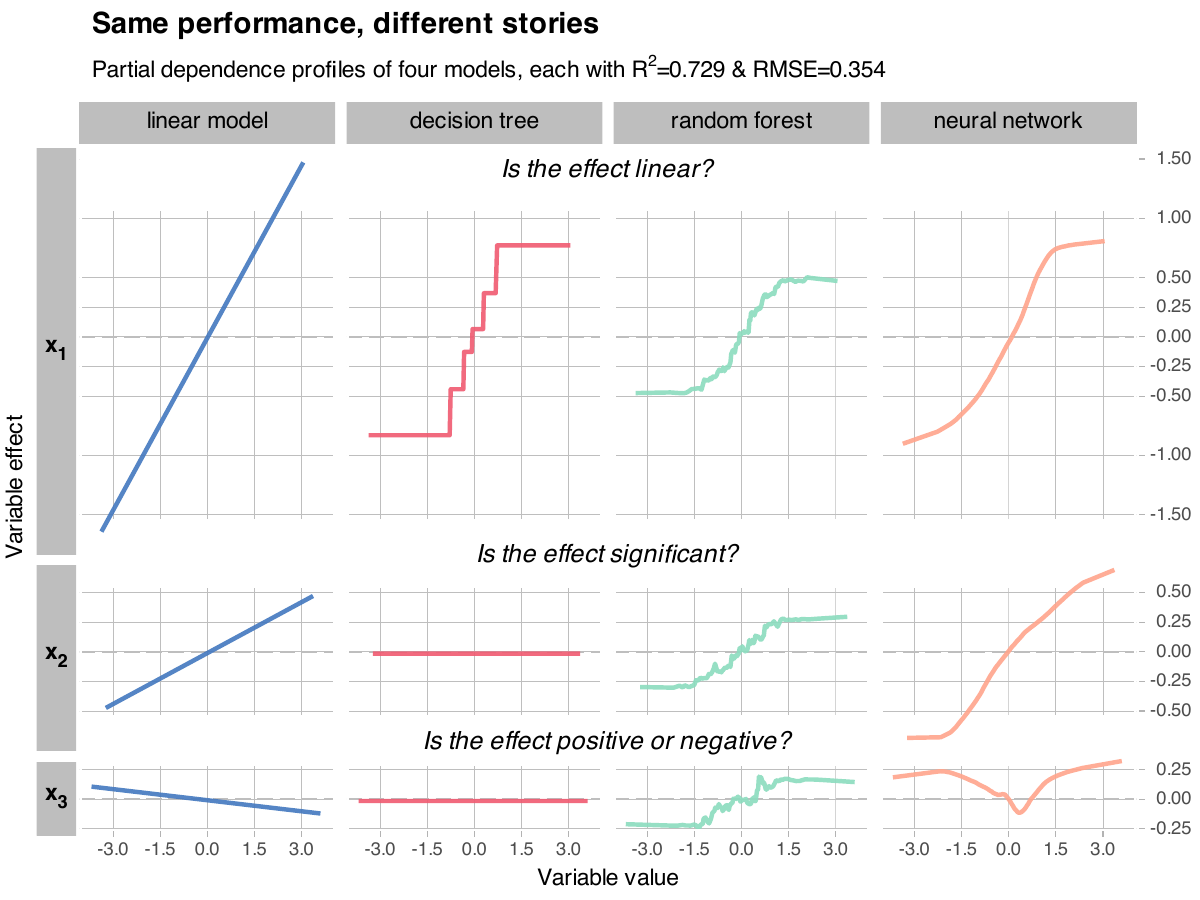}
\end{center}
\caption{Overview of the Rashomon Quartet: four models of different types -- linear model, decision tree, neural network, and random forest -- with the same predictive performance ($R^2=0.729$, $RMSE=0.354$) but different behaviors. Each panel shows partial dependence profiles for the three variables $x_1$, $x_2$, and $x_3$. All models agree that $x_1$ is strongly linked with $y$ but disagree on whether the relation is linear. The models disagree on how variables $x_2$ and $x_3$ are related to $y$. 
}\label{fig:quartet}
\end{figure}


To go beyond predictive performance and compare explanations provided by various predictive models, we introduce the \emph{Rashomon Quartet}, shown in Figure \ref{fig:quartet}. 
These are four predictive models for a regression task fitted to a manually crafted synthetic dataset. The data has three explanatory variables and 1000 observations. Model 1 is a linear model, model 2 is a decision tree, model 3 is a random forest, while model 4 is a neural network. All four models achieve equally good performance with $R^2 = 0.729$ and $RMSE = 0.354$ on the separate test set. Random seeds and input parameters for models 2-4 were controlled to ensure these statistics agreed at the third and fourth decimal places (Appendix D). 

A model's behavior can be examined using partial dependence (PD) profiles~\citep{Friedman00greedyfunction}, which generalize main effect plots~\citep{JSSv008i15} typically used for linear models to a broader class of predictive models. A PD profile for variable $x_i$ and model $f$ is the model response averaged over the distribution of the remaining variables, i.e.  
$PD(x_i;f) = \int f(x_i,x_{-i})p(x_{-i})dx_{-i}$, where $x_{-i}$ stands for vector of all variables except $x_i$ while $p(x_{-i})$ stands for its marginal density. 

Figure \ref{fig:quartet} shows PD profiles for each model and explanatory variable from the Rashomon Quartet. The differences between models are clear. The decision tree uses only the $x_1$, but the linear model is built on $x_1$ and $x_2$, and a small negative contribution from $x_3$. The random forest uses all three variables, with a smaller positive contribution from $x_3$. Contrastingly, the neural network uses a non-monotonic relationship with $x_3$. 

\begin{figure}[!ht]
\begin{center}
\includegraphics[width=\textwidth]{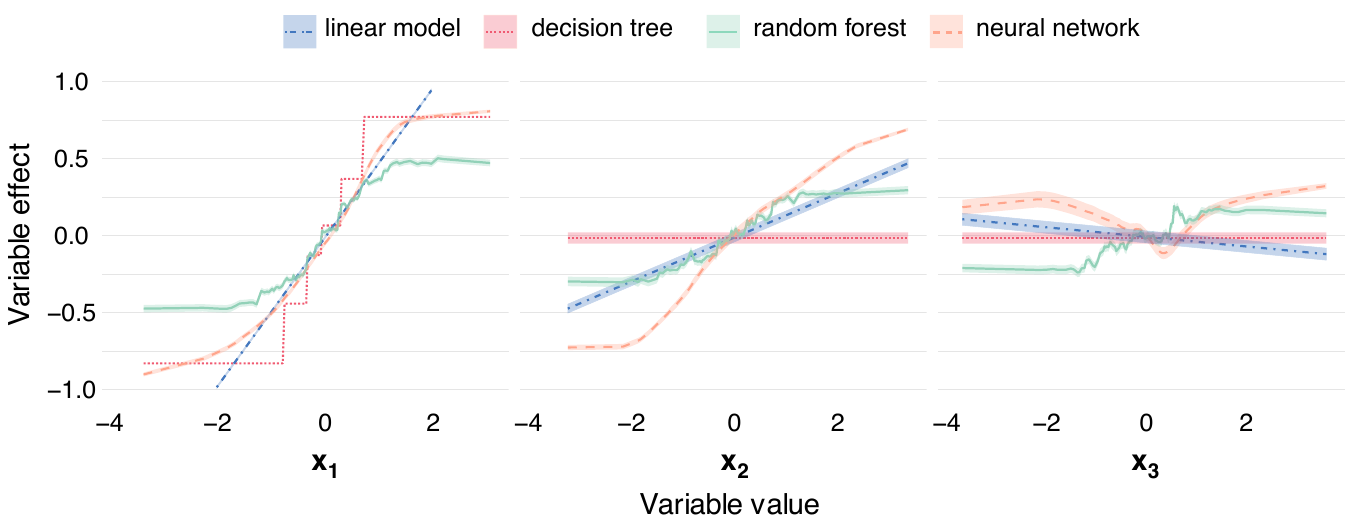}
\end{center}
\caption{Partial dependence profiles with point-wise 95\%-confidence intervals shown by line thickness. There is very little overlap between intervals which suggests that the models are confidently different in their view of the relationship with $y$.}\label{fig:conf-intervals}
\end{figure}

Figure~\ref{fig:conf-intervals} adds confidence intervals using line thickness to the PD profiles, computed using the nonparametric bootstrap sampling following the method in \cite{molnar2021relating}. These intervals do not overlap for most values of the predictors, providing support for concluding that there are differences in the model fits. 

In order to build an example clearly exhibiting this phenomenon, a synthetic dataset with three correlated input variables $x_1, x_2, x_3$ and target variable $y$ was simulated from:

\begin{equation}
y = \sin\left(\left(3 x_1 +  x_2\right)/5\right) + \varepsilon,
\label{eq:yformula}
\end{equation}

\noindent where $\varepsilon \sim \mathcal N (0, \frac 13)$ while $[x_1, x_2, x_3] \sim \mathcal N(0, \Sigma_{3\times 3})$ and $\Sigma_{3\times 3}$ has $1$ on the diagonal and $0.9$ beyond diagonal. Note that $x_3$ is not present in the equation (\ref{eq:yformula}) but is correlated with the other predictors. A scatterplot matrix of the data comparing training and test sets can be found in Appendix E.

The data was structured to exploit the design decisions of each class of models to reach an $R^2$ close to $0.7$ while interpreting the data in different ways.

\textbf{The story told by the linear model\footnote{Linear models are a very versatile family; here we consider only its simplest (default) subset of models with a linear relationship between $x_i$ and $y$.}} (first column in Figure \ref{fig:quartet}).
Analysis of the fitted linear model suggests that $x_1$ is the most important variable.  Its coefficient is more than three times larger than that of $x_2$. The coefficient on $x_3$ is the smallest and negative (though not statistically significant). A more detailed analysis of the residuals would suggest more flexible models, such as polynomial regression, but in order to see this, one cannot just look at performance but needs to make residual plots.

\textbf{The story told by the decision tree\footnote{For illustrative purposes, we have limited ourselves to trees with a maximum depth of 3.}} (second column in Figure \ref{fig:quartet}). Analysis of the fitted tree model suggests that the only important variable is $x_1$. This differs from the story told through the linear model, not only by the absence of $x_2$, but also from different behavior on the extremes $x_1$, where it plateaus. Why has this happened? In successive splits, the variables are chosen greedily, so it may happen that the effect of $x_2$ is ignored because the tree will choose the correlated dominant $x_1$, exhausting the splits available to the decision tree. 

\textbf{The story told by the random forest} (third column in Figure \ref{fig:quartet}). All three variables are important, $x_1$ twice as strong as $x_2$ and $x_3$. The shape and nature of the relationship for each variable is similarly positive.
Why is $x_3$ more prominent in this model fit?  This is due to the random sampling of variables for each tree in the random forest fit. Because $x_3$ is correlated with $x_1$, it will contribute similarly to the response value.

\textbf{The story told by the neural network\footnote{Consistently obtaining results in neural network fits is still hard, so this fit was ensured by controlling the random number seed and parameterization.}}  (fourth column in Figure \ref{fig:quartet}). For a neural network, the variables $x_1$ and $x_2$ are equally important and have a monotonic effect on $y$. The effect of the $x_2$ variable is stronger than in the other three model fits. Curiously, $x_3$ has a non-monotonic relationship with $y$.


\textbf{Conclusions} Anscombe wrote: \textit{This article is emphatically not a catalog of useful graphical procedures in statistics. Its purpose is merely to suggest that graphical procedures are useful}. His ideas guide our work, to provide a simple set of models which all tell different stories. We hope that the Rashomon Quartet will be useful for teaching, and for testing exploratory methods for understanding and comparing models. It hopefully motivates the development of more examples of Rashomon quartets to help with the understanding of predictive models in general.  Today, performance is not enough.

Although the Rashomon Quartet models tell different stories, several aspects should be kept in mind. Correlated variables provide a simple framework on which to produce different fits. These relationships can be better examined using marginal profiles, or accumulated local effects~\citep{apley2020visualizing}. 
The four models of three predictors were carefully chosen, a simpler scenario containing two models and single predictor can also be used (see the Rashomon Couple in Appendix A).


\section*{Supplementary materials}

Detailed analysis, datasets, models, and source code are available at \url{https://github.com/MI2DataLab/rashomon-quartet}. 

\section*{Acknowledgements}

We appreciate the careful reviews provided by two anonymous reviewers,  who discussed in great detail the weaknesses of the presented models and the potential for better ways to tell the story of the Rashomon Quartet. The first three authors contributions were financially supported by SONATA BIS grant 2019/34/E/ST6/00052 from the Polish National Science Centre (NCN). The results and plots in the paper were produced using these R packages: DALEX~\citep{DALEX}, partykit~\citep{partykit}, randomForest~\citep{randomForest}, neuralnet~\citep{neuralnet}, ggplot2~\citep{ggplot2}, GGally~\citep{GGally}.

\clearpage
\bibliographystyle{jasa3}
{\small 
\bibliography{references}
}

\clearpage

\bigskip
\begin{center}
{\large\bf SUPPLEMENTARY MATERIAL}
\end{center}

\section*{Appendix A: Rashomon Couple}

The concept behind the Rashomon Quartet can be presented on analytical grounds. The easiest way to do this will be for two very popular families of models (linear and tree regression models) of the same complexity (parameterized by a single parameter). The following example is constructed based on the analysis of $|x|^\alpha$ functions on the interval $x \in [-1, 1]$. To make this paper concise, we show the result for a single function from this family. The reader can easily derive the estimators and prediction errors for other functions in this class.

Consider a data generation function 
\begin{equation}
  f(x) = sign(x){\lvert x \rvert}^{\frac{\sqrt{3}-1}{2}}  
\end{equation}
with $x \sim \mathcal U[-1,1]$.

Now consider a family $\mathcal F_{LM}(b_1)$ of linear models of the form $y = b_1 x$. It can be easily shown that the OLS estimator\footnote{For simplicity, we are working here with expected values that correspond to limits with the sample size increasing to infinity.} for $b_1$ is 
\begin{equation}
  \hat b_1 = \frac{6}{\sqrt 3 + 3},
\end{equation}
so the best model, according to the OLS error rate, is
\begin{equation}
  \hat f_1(x) = \frac{6}{\sqrt 3 + 3} x,
\end{equation}
while the corresponding mean squared error (MSE) rate is
\begin{equation}
  MSE(\hat f_1) = \frac 1{\sqrt 3} - \frac {12}{({\sqrt 3 + 3})^2}.
\end{equation}

Now consider another family $\mathcal F_{BT}(b_0)$ of shallow binary decision tree models of the form $y = sign(x)\cdot b_0$. It can be easily shown that the OLS estimator for $b_0$ is 
\begin{equation}
  \hat b_0 = \frac{2}{\sqrt 3 + 1},
\end{equation}
so, the best model from this family, according to the OLS error rate, is
\begin{equation}
  \hat f_0(x) =  sign (x) \frac{2}{\sqrt 3 + 1},
\end{equation}
while the corresponding MSE rate is
\begin{equation}
  MSE(\hat f_0) = \frac 1{\sqrt 3} - \frac {12}{({\sqrt 3 + 3})^2}.
\end{equation}

\begin{figure}[b!]
\begin{center}
\includegraphics[width=0.5\textwidth]{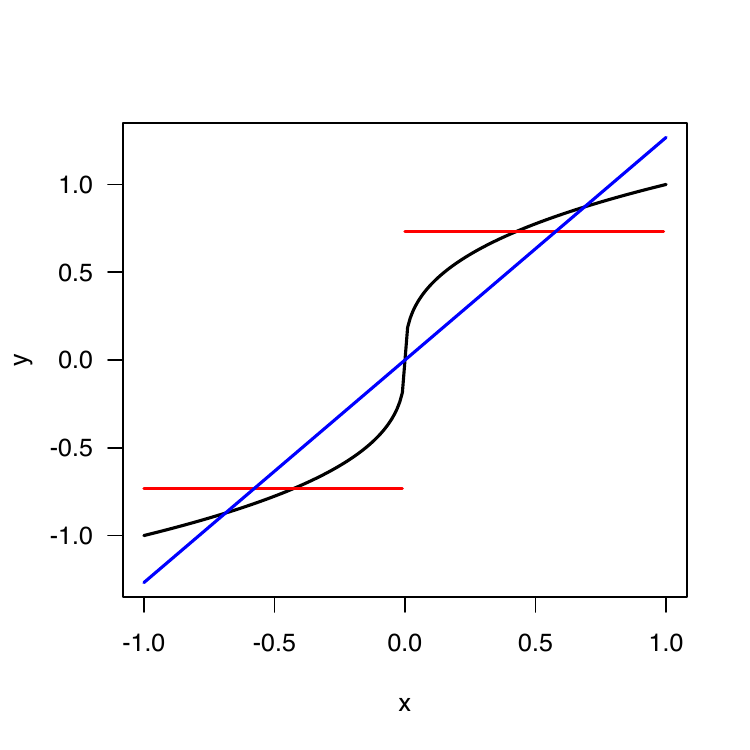}
\end{center}
\caption{Example of a Rashomon couple. The black curve corresponds to the true data generating function $f(x) = sign(x){\lvert x \rvert}^{\frac{\sqrt{3}-1}{2}}$, while the blue curve corresponds to the best model against the OLS criterion in the family $\mathcal F_{LM}(b_1)$ of linear functions, and the red curve in the family $\mathcal F_{BT}(b_0)$ of shallow binary trees.}\label{fig:example}
\end{figure}

Note that $MSE(\hat f_1) = MSE(\hat f_0)$ event though $\hat f_1 \neq \hat f_0$. Figure \ref{fig:example} shows the function $f(x)$ and the estimated best models in families $\mathcal F_{LM}(b_1)$ and $\mathcal F_{BT}(b_0)$. Both models have exactly the same loss function.

This simple observation can be easily extended to higher dimensions. The Rashomon Quartet is an example that this situation is not just theoretically possible, but will also happen (even can be engineered) to a larger number of model families in higher dimensions.

\clearpage

\section*{Appendix B: Analysis of model residuals}

Partial dependence profiles are not the only technique for explaining model behavior. Residual analysis can also be used for the purpose of model comparison. 

\begin{figure}[b!h]
\begin{center}
\includegraphics[width=0.8\textwidth]{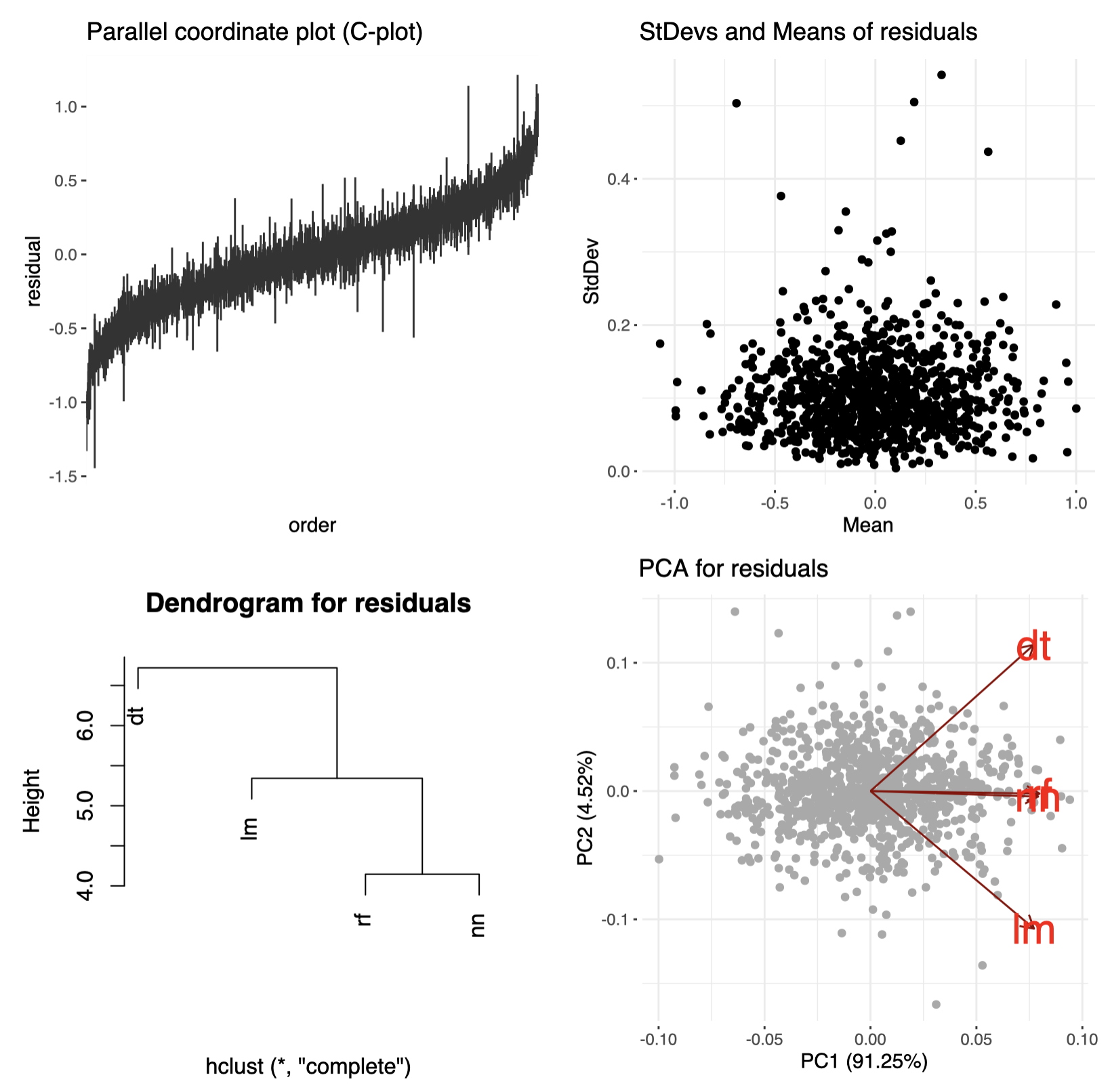}
\end{center}
\caption{The parallel coordinate plot depicts ranges for residuals for different models, one range per observation ordered along the mean value. The second panel shows the difference between model averages and standard deviations for residuals, one point per observation. The following panels show the dendrogram and PCA for residuals.}\label{fig:residuals2}
\end{figure}

\begin{figure}[b!h]
\begin{center}
\includegraphics[width=0.65\textwidth]{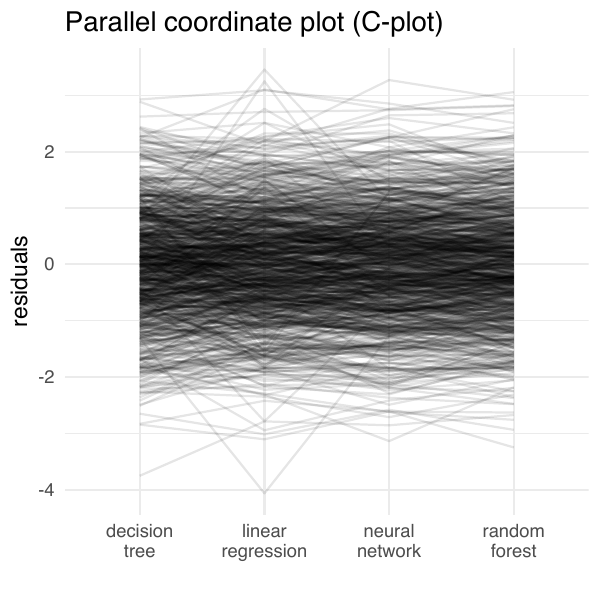}
\end{center}
\caption{The parallel coordinate plot of residuals for the Rashomon Quartet models. Lines connect the residuals for the same observations. Most of the lines are flattened, suggesting a significant correlation of residuals across models.}\label{fig:residuals}
\end{figure}

In Figures \ref{fig:residuals2} and \ref{fig:residuals}, we show a set of residuals analysis inspired by the work of \cite{UNWIN2003553}. 
It shows that the models' residuals are correlated, so the models similarly underestimate or overestimate the data, which may result from large noise in the target variable. 

\clearpage

\section*{Appendix C: How to engineer your own Rashomon quartet}

In the main body of this paper, we showed that models from a Rashomon quartet have different behavior but very similar performances. Yet, the data on which the models are trained looks like it was pulled out of a magical statistical hat. In this appendix, we explain how to construct the data to increase the chance of models with such properties appearing.

Models in a Rashomon set can come from one family of models or from different families. In this work, we assumed that it would be more spectacular to find representatives from different families. So, the first step is to select the families of models for which we want to find elements of a Rashomon set. The selection of simple linear models was obvious because it is the most popular class of regression models. The choice of decision tree models was natural because they learn a different nature of relationships than linear models (interactions instead of additive relationships). The choice of random forest and neural network models was due to their popularity in the machine learning community.

The next step is to find a family of data generation functions that are composed of different data patterns. The data patterns are selected to manipulate MSE specifically for the selected family of models. To do this, it is necessary to have an intuition of how model families behave in the presence of certain data patterns. Table \ref{tab:bingo} summarises intuitions associated with several common data patterns for selected families.

The first data pattern to optimize is nonlinearity. For a well-specified model, a linear model will have the lowest MSE, so to equalize MSE for linear models and other families, it is necessary to use a transformation in which the level of nonlinearity can be adjusted. On the interval $[-1,1]$ such a transformation is $\sin(\theta_1 x)$. The larger $\theta_1$, the greater the linearity. 

Having control over the relative performance of a linear model, another data pattern is chosen to differentiate the performance of models from different families. The nonlinearity of the $\sin$ function is most pronounced at the edges of the domain of $x$. To differentiate the learnability of this signal, it is sufficient to manipulate the mass of the distribution in the edges from which $x$ is sampled. Neural networks will approximate this signal better than tree models. We considered various parameterized families of distributions like Beta (two parameters) of t-Student (one parameter) but the obtained results were too complex. So, in the end, we used normal distribution for sampling $x$.

The models generated in this way had quite similar behavior. To further differentiate them, another data pattern was used, which is the correlation between variables combined with the fact nonlinear function is applied to a linear combination of two predictors. The correlated variables are utilized differently by a linear model (the coefficients may cancel each other out) than by a random forest model. 
Thus, we have the following family of data-generating functions 
\begin{equation}
y = \sin\left(\theta_1 (x_1 +  \theta_2 x_2)\right) + \varepsilon,
\label{eq:yformula2}
\end{equation}

where $[x_1, x_2, x_3] \sim \mathcal N(0, \Sigma_{3\times 3})$ and $\Sigma_{3\times 3}$ has $1$ on the diagonal and $\theta_3$ beyond diagonal. To control relative MSE parameters $\theta_2$ and $\theta_3$ need to be optimized. 

At this stage, the models trained on such data had different behavior. We manually performed a search of good $\theta$ parameters looking for reasonably round values that lead to the similar performance of models. Finding a good configuration requires about a few dozen iterations to get  MSE equalized at the second decimal place. The final step was to optimize the random seed for data generation to equalize the performance of the trained models further, that is, to be consistent at the third digit and very similar at the fourth digit. This required reviewing about a thousand different seeds for the training data. The test set was not optimized, it is large enough to get a stable estimate of MSE.

The fewer the families of models, the easier it is to balance their performance. In the case of the \emph{Rashomon couple} (see Appendix A), it was enough to consider a family of functions $x^\theta$ where $\theta$ adjusts the level of linearity. Then, all we had to do was select two model families with different behavior (decision trees and linear models, each with a single coefficient, $b_0$ and $b_1$ respectively), determine the performance as functions of $\theta$ and choose $\theta$ in such a way as to balance the performance of the model from both families.

\begin{table}[h!]
\footnotesize
    \centering
\begin{tabular}{p{3cm}|p{3cm}|p{3cm}|p{3cm}|p{3cm}}
\backslashbox[3.5cm]{\textbf{data patterns}}{\textbf{model family}} & linear models & neural networks & decision tree & random forest \\ \hline
nonlinear relation between $x_i$ and $y$ & without transformation of variables, linear models will not learn nonlinear relationships; the higher the nonlinearity, the \textbf{higher MSE} & can learn smooth nonlinear relationships, \textbf{opportunity for lower MSE} & able to learn stepwise relationships. \textbf{Low or high MSE possible} depending on the data generation function & able to learn a very wide class of relationships \\ \hline
light tails in $x_i$   & \textbf{extreme observations may be influential}, have high leverage and shift $\beta$ estimates in a specific direction & \textbf{extreme observations may be very influential}, large variance at the edges is possible because the residuals for observations in tails are rare and have small impact on MSE & tendency for \textbf{flat predictions at edges}, especially if the hyperparameter \textit{minimum leaf size} is high & similar to decision trees \\ \hline
highly correlated predictors $x_i$ and $x_j$ & chance for negatively correlated coefficients $\beta_i$ and $\beta_j$ \textbf{compensating each others} & \centering --- & chance that \textbf{only one variable will be used} because the other does not bring gain in MSE  & drawing a subset of variables for each split will cause the correlated variables to be used similarly in different trees \\ \hline
 interactions between $x_i,x_j$ and $y$ & if explicitly not included in the model then they will not be learned. The higher the interactions, the \textbf{higher MSE} & able to learn & able to learn & able to learn, \textbf{but} due to the sampling of a subset of variables, some trees may not have the chance to see both variables \\ 
\end{tabular}
    \caption{Expected model behavior matrix. What to expect if certain patterns occur in the training data}
    \label{tab:bingo}
\end{table}

\clearpage

\section*{Appendix D: Reproducibility of the results}

All the code snippets necessary to reproduce the data, models, and visualizations can be found at \url{https://github.com/MI2DataLab/rashomon-quartet}. For convenience, we place these code snippets below. The necessary files are available in the GitHub repository.

\subsection*{Read data}

\begin{verbatim}
train <- read.table("rq_train.csv", sep=";", header=TRUE)
test  <- read.table("rq_test.csv",  sep=";", header=TRUE)
\end{verbatim}

\subsection*{Train models}

\begin{verbatim}
set.seed(1568) 
library("DALEX")
library("partykit")
model_dt <- ctree(y~., data=train, 
                  control=ctree_control(maxdepth=3, minsplit=250))
exp_dt <- DALEX::explain(model_dt, data=test[,-1], y=test[,1], 
                         verbose=F, label="decision tree")
mp_dt <- model_performance(exp_dt)
imp_dt <- model_parts(exp_dt, N=NULL, B=1)

model_lm <- lm(y~., data=train)
exp_lm <- DALEX::explain(model_lm, data=test[,-1], y=test[,1], 
                         verbose=F, label="linear regression")
mp_lm <- model_performance(exp_lm)
imp_lm <- model_parts(exp_lm, N=NULL, B=1)

library("randomForest")
model_rf <- randomForest(y~., data=train, ntree=100)
exp_rf <- DALEX::explain(model_rf, data=test[,-1], y=test[,1], 
                         verbose=F, label="random forest")
mp_rf <- model_performance(exp_rf)
imp_rf <- model_parts(exp_rf, N=NULL, B=1)

library("neuralnet")
model_nn <- neuralnet(y~., data=train, hidden=c(8, 4), threshold=0.05)
exp_nn <- DALEX::explain(model_nn, data=test[,-1], y=test[,1], 
                         verbose=F, label="neural network")
mp_nn <- model_performance(exp_nn)
imp_nn <- model_parts(exp_nn, N=NULL, B=1)
\end{verbatim}

\subsection*{Calculate performance}

\begin{verbatim}
mp_all <- list(lm=mp_lm, dt=mp_dt, nn=mp_nn, rf=mp_rf)
R2   <- sapply(mp_all, function(x) x$measures$r2)
round(R2, 4)
#     lm     dt     nn     rf 
# 0.7290 0.7287 0.7290 0.7287 

rmse <- sapply(mp_all, function(x) x$measures$rmse)
round(rmse, 4)
#     lm     dt     nn     rf 
# 0.3535 0.3537 0.3535 0.3537
\end{verbatim}

\subsection*{Visualize models}

\begin{verbatim}
plot(model_dt)
summary(model_lm)
model_rf
plot(model_nn)
\end{verbatim}

\subsection*{Visualize variable importance}

\begin{verbatim}
plot(imp_dt, imp_nn, imp_rf, imp_lm)
\end{verbatim}

\subsection*{Visualize partial dependence profiles}

\begin{verbatim}
pd_dt <- model_profile(exp_dt, N=NULL)
pd_rf <- model_profile(exp_rf, N=NULL)
pd_lm <- model_profile(exp_lm, N=NULL)
pd_nn <- model_profile(exp_nn, N=NULL)

plot(pd_dt, pd_nn, pd_rf, pd_lm)
\end{verbatim}

\subsection*{Plot data distribution}

\begin{verbatim}
library("GGally")
both <- rbind(data.frame(train, label="train"),
              data.frame(test, label="test"))
ggpairs(both, aes(color=label),
        lower = list(continuous = wrap("points", alpha=0.2, size=1), 
                     combo = wrap("facethist", bins=25)),
        diag = list(continuous = wrap("densityDiag", alpha=0.5, bw="SJ"), 
                    discrete = "barDiag"),
        upper = list(continuous = wrap("cor", stars=FALSE))) 
\end{verbatim}

\clearpage

\section*{Appendix E: Variable distributions}

\begin{figure}[!htb]
\begin{center}
\includegraphics[width=\textwidth]{
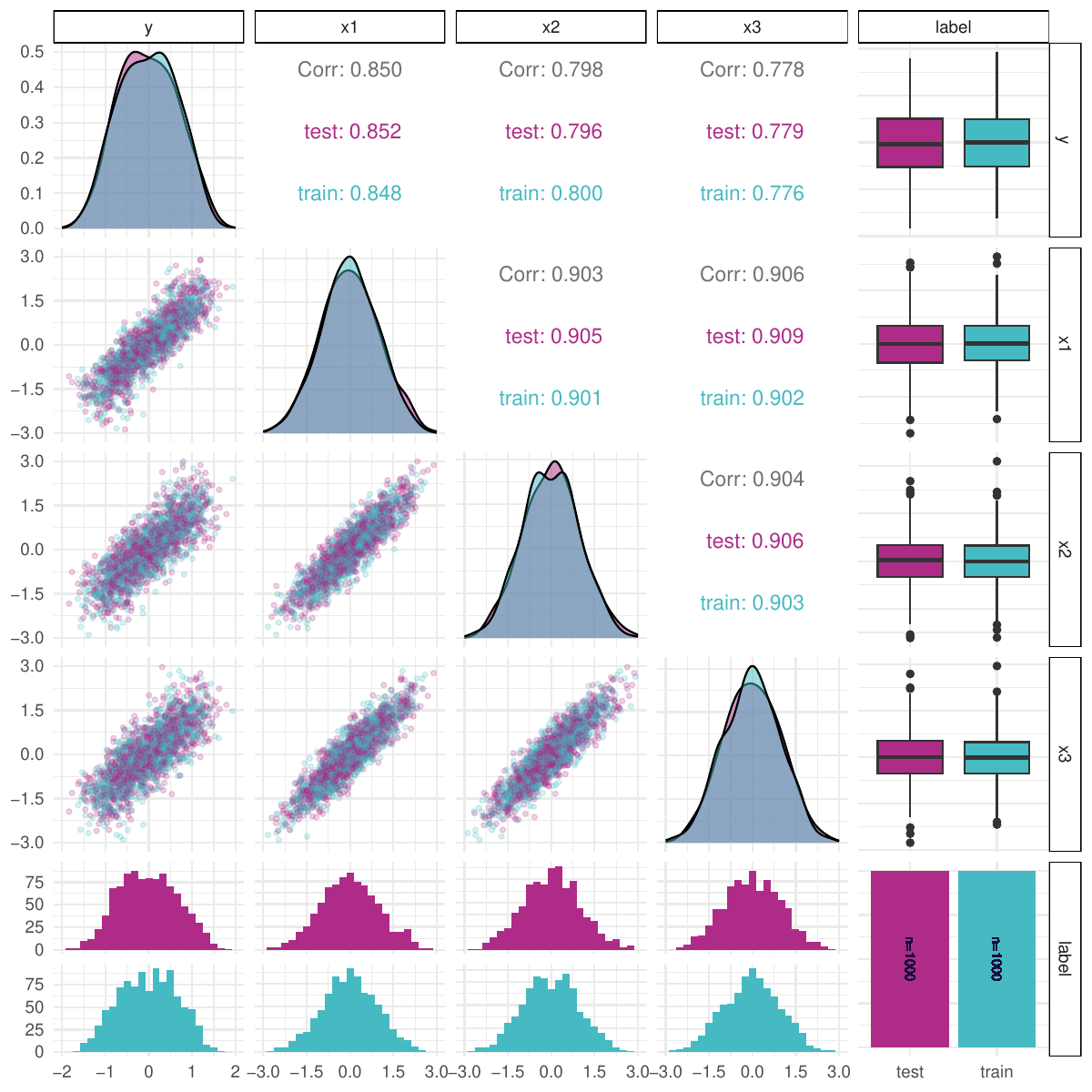}
\end{center}
\caption{Visualization of the Rashomon Quartet datasets (train and test sets). Variables $x_1$, $x_2$ and $x_3$ follow normal distributions $\mathcal N(0,1)$ and are correlated. Variables' distribution between the train and test set are the same.}
\label{fig:data}
\end{figure}

\end{document}